\documentclass[11pt]{article}

\usepackage[final]{acl}

\usepackage{times}
\usepackage{latexsym}

\usepackage[T1]{fontenc}

\usepackage[utf8]{inputenc}

\usepackage{microtype}

\usepackage{inconsolata}

\usepackage{graphicx}
\usepackage{cuted}

\makeatletter
\def\input@path{{../}{../text/}{../tables/}{../imgs/}}
\makeatother
\graphicspath{{../}{../imgs/}}

\usepackage{balance} 

\usepackage{algorithm}
\usepackage{algorithmic}

\usepackage{amssymb} 
\usepackage{amsmath}
\usepackage{enumitem}
\usepackage[utf8]{inputenc} 
\usepackage{booktabs}       
\usepackage{amsfonts}       
\usepackage{nicefrac}       
\usepackage{microtype}      
\usepackage{multirow}       
\usepackage{xspace}         
\usepackage{listings}       
\usepackage{soul}           
\usepackage[table]{xcolor} 

\newcommand{\eg}{\textit{e.g.}}
\newcommand{\ie}{\textit{i.e.}}

\title{ToolPRM: Fine-Grained Inference Scaling of Structured Outputs for Function Calling}

\author{
  \textbf{Jianghao Lin\textsuperscript{1}\thanks{Equal contribution.}},
  \textbf{Yuanyuan Shi\textsuperscript{1}\footnotemark[1]},
  \textbf{Xin Peng\textsuperscript{2}},
  \textbf{Renjie Ding\textsuperscript{2}},
  \textbf{Hairui Wang\textsuperscript{2}},
\\
  \textbf{Yuxuan Peng\textsuperscript{2}},
  \textbf{Bizhe Bai\textsuperscript{3}},
  \textbf{Weixi Song\textsuperscript{3}},
  \textbf{Fengshuo Bai\textsuperscript{1}},
\\
  \textbf{Huacan Chai\textsuperscript{1}},
  \textbf{Weinan Zhang\textsuperscript{1,3}\thanks{Corresponding authors.}},
  \textbf{Fei Huang\textsuperscript{2}\footnotemark[2]},
  \textbf{Ying Wen\textsuperscript{1,3}\footnotemark[2]}
\\
  \textsuperscript{1}Shanghai Jiao Tong University, China;
\\
  \textsuperscript{2}Longshine AI Research, China;
\\
  \textsuperscript{3}Shanghai Innovation Institute, China
\\
  \texttt{\{linjianghao, wnzhang, ying.wen\}@sjtu.edu.cn, \href{mailto:huangfei@longshine.com}{huangfei@longshine.com}}
}

\begin{document}
\maketitle

\begin{abstract}
Large language models (LLMs) excel at function calling, but inference scaling has been explored mainly for unstructured generation. We propose an inference-scaling framework for structured outputs that combines fine-grained beam search with \textbf{ToolPRM}, a process reward model scoring each intra-call decision (function name and argument filling). We build the first fine-grained intra-call supervision dataset via function masking, rollout collection, and step-level annotation. ToolPRM outperforms outcome and coarse-grained reward models in predictive accuracy and yields consistent test-time gains on multiple function-calling benchmarks. We further show that structured generation follows ``\textbf{explore more but retain less}'', since early JSON errors are unrecoverable. 
\end{abstract}

\section{Introduction}
\label{sec:intro}

Large language models (LLMs) have demonstrated remarkable capabilities in a diverse range of tasks~\cite{zhou2026externalization,xi2025survey}. Subsequently, inference scaling has emerged as a critical technique for further enhancing their performance, where additional computational resources are allocated during the inference phase to explore a wider range of possibilities or to iteratively refine outputs, and to ultimately maximize the quality of the output. The core idea behind inference scaling strategies is to move beyond generating a single greedily decoded sequence of tokens. Instead, inference scaling approaches often intentionally increase the computational effort at inference time, to search for a wider range of possibilities and multiple potential reasoning paths, and select the best trajectories for better performance~\cite{ke2025surveyfrontiersllmreasoning}. 

Existing research on inference scaling predominantly focuses on \textbf{unstructured output generation tasks}, such as mathematical problems~\cite{puri2025probabilisticinferenceapproachinferencetime} and other intricate reasoning tasks~\cite{ma2023letsrewardstepstep}, where each sentence or discrete unit serves as a processing step. Novel methods often combine Tree of thoughts (ToT)~\cite{yao2023treethoughtsdeliberateproblem} with tree-based search algorithms, generalizing the linear Chain-of-Thought~\cite{wei2022chain} (CoT) by collecting multiple potential next thoughts from a given state. To search for the optimal output while condensing the possibility space, beam search and Monte Carlo Tree Search (MCTS) become the ideal choices, and several studies have adapted them for LLM inference scaling~\cite{wu2025inferencescalinglawsempirical, zhang2023planninglargelanguagemodels, liu2023don, choi2023kcts, zhou2024languageagenttreesearch}. Correspondingly, they involve either self-evaluation~\cite{xie2023selfevaluationguidedbeamsearch} or process reward models~\cite{ma2023letsrewardstepstep, hung2025rewardGuidedTreeSearch} (PRMs) to generate an evaluation score for each sub-sequence, as the reference for searching. However, the application of inference scaling techniques within the domain of \textbf{structured outputs}, particularly in the context of function calling\footnote{In this paper, the term ``function'' is interchangeable with ``tool'' and ``API''.}, remains significantly underexplored. 

Current studies that incorporate process reward mechanisms and inference scaling for function calling tasks operate at a coarse-grained level~\cite{wang2024fantasticsequencesthemfaithful,nath2025toolcomp}, treating an entire round of function calling response as a singular, monolithic step. For example,  Wang et al.~\cite{wang2024fantasticsequencesthemfaithful} employ a Best-of-N approach, where multiple generated function call candidates are scored and ranked by an outcome reward model (ORM). 
The final answer is selected by assessing the entire generation as a singular entity without applying distinct fine-grained rewards to its sub-components within each function call. 
This overlooks the inherent multi-stage nature of the function calling process that involves solving a sequence of sub-questions, including the selection of function name, the identification of relevant parameters, and the decision of parameter values. Consequently, the potential for optimization of fine-grained process rewards at the intra-call level is not adequately addressed by existing methodologies. 

To this end, in this paper, we propose a novel fine-grained process reward mechanism specifically designed for structured function calling tasks (dubbed \textbf{ToolPRM}). 
In contrast to prior methods that treat function calls as indivisible units, our approach decomposes each call into semantically interpretable intermediate steps. 
We develop a dedicated process reward model, ToolPRM, supervised using meticulously curated fine-grained intra-call step labels constructed from both xlam-function-calling-60k~\cite{liu2024apigen} and xlam-irrelevance-7.5k~\cite{lin2024hammerrobustfunctioncallingondevice} datasets. 
Consequently, we develop fine-grained beam search with the intra-call process supervision from ToolPRM, which proves to  outperform existing function-calling baselines in our experiments. 

More critically, we highlight an important insight for the inference scaling on structured and unstructured output generation. 
In traditional inference scaling settings like math reasoning, the unstructured outputs usually enable more flexible inference scaling by simply maintaining a larger number of candidates. 
Even if we maintain the wrong intermediate steps, it could be further corrected at later steps via reflection and process supervision.
However, in structured output generation, retaining multiple candidate trajectories often degrades performance, as errors in early steps cannot be corrected later, thereby wasting the subsequent computational budget. 
Therefore, we propose a tailored inference scaling principle for structured outputs: devote computation to exploring a wider range of decisions (i.e., increasing the beam width), while aggressively pruning the number of retained candidates (i.e., reducing the number of active beams). 
We summarize such a principle of inference scaling for structured function calling outputs as ``\textbf{explore more, but retain less}''.

The contributions of this paper are as follows:
\begin{itemize}[leftmargin=10pt]
    \item \textbf{Fine-Grained Intra-Call Process Supervision Dataset}: We construct a novel annotated dataset specifically designed for fine-grained intra-call reward modeling of structured function calling. 
    This dataset, along with its annotation scripts, is given in the anonymous link and will be made publicly available, facilitating future research and benchmarking in fine-grained process supervision for structured output generation. 

    \item \textbf{ToolPRM}: We introduce ToolPRM, a fine-grained process reward modeling framework specifically tailored for inference scaling in structured function calling tasks. ToolPRM could assist different backbone models in conducting test-time scaling for superior performance.

    \item \textbf{Inference Scaling Principle for Structured Outputs}: 
    We identify and formalize a critical principle for inference scaling in structured output tasks: \textit{Explore more but retain less}. It highlights the importance of widening exploration while aggressively eliminating incorrect, unrecoverable steps, leading to better performance.
\end{itemize}

\section{Related Works}
\label{sec: related work}

\paragraph{LLM for Function Calling.} Recent research has increasingly demonstrated the significant potential of enabling large language models (LLMs) to interact with external systems through function calling~\cite{liu2024apigen,park2025flexibleefficientgrammarconstraineddecoding,srinivasan2023nexusraven,chai2025parl,yang2025survey,qu2025tool}. 
For instance, IBM's Granite-20B-FunctionCalling~\cite{abdelaziz2024granite} enhances performance through multi-task learning on seven core function-calling subtasks. 
ToolACE~\cite{liu2024toolace} designs a self-evolution synthesis process for high-quality complex function calling data generation. 
To further enhance the LLM response to complex function calling, in this paper, we disassemble function calling into fine-grained basic steps, employ a process reward model to better guide the generation steps inside function calling, and finally implement inference scaling mechanisms with intra-call-level granularity.

\paragraph{Inference Scaling Strategies.} A variety of inference scaling strategies have been developed based on existing sampling and searching algorithms. 
Straightforward methods include self-consistency~\cite{wang2023selfconsistencyimproveschainthought} strategy, which samples N independent candidate reasoning paths through temperature sampling~\cite{ACKLEY1985147} and decides the ultimate answer with majority voting.
Best-of-N (BoN)~\cite{brown2024largelanguagemonkeysscaling} evaluates the N candidates with an outcome reward model to select the best one. 
In novel studies, tree-based methods have become the mainstream, and effective algorithms based on beam search or Monte Carlo Tree Search (MCTS) emerge~\cite{wu2025inferencescalinglawsempirical,zhang2023planninglargelanguagemodels, liu2023don, choi2023kcts, zhou2024languageagenttreesearch}. 
While substantial studies have verified the effectiveness of inference scaling strategies on unstructured output generation tasks, especially on mathematical reasoning tasks, the application on structured output generation tasks remains underexplored. 
In this paper, we apply inference scaling on structured function call generation, analyze its promotion on final performance, and discuss the balance between exploration and retention in searching process.

\paragraph{Process Reward Models (PRMs).} In complex multi-step reasoning tasks where errors can occur at any point, granular feedback is particularly valuable~\cite{zhou2025steporlm,zheng2025survey}. In contrast to the outcome reward models (ORMs) that just evaluate the final outcome, PRMs are trained to evaluate the correctness or quality of intermediate steps within an inference process~\cite{uesato2022solvingmathwordproblems,zheng2025cold,zhu2025retrieval}. 
PRMs are initially applied in RLHF~\cite{ouyang2022traininglanguagemodelsfollow} training process to provide reward signals and to supervise the generation of reasoning steps~\cite{piotrowski2025lightweightlatentverifiersefficient,lightman2023letsverifystepstep}.
In this work, we curate the first fine-grained intra-call  process supervision dataset with automatic annotation of granular process rewards, and finetune the ToolPRM that excels at verifying the intermediate steps in forming a full function call. Combined with well-configured searching strategies, our proposed ToolPRM helps the function-calling LLM reach the state-of-the-art performance.
\section{Methodology}
\label{sec:methodology}

This section introduces the fine-grained process reward model for function calling (\ie, ToolPRM). We first present fine-grained decomposition for each turn of function callings, and present the data collection and annotation for reward modeling training. Then, we design a state transition machenism for function calling, and apply beam search with fine-grained process supervision with ToolPRM.

\begin{figure*}[t]
    \centering
    \includegraphics[width=0.99\textwidth]{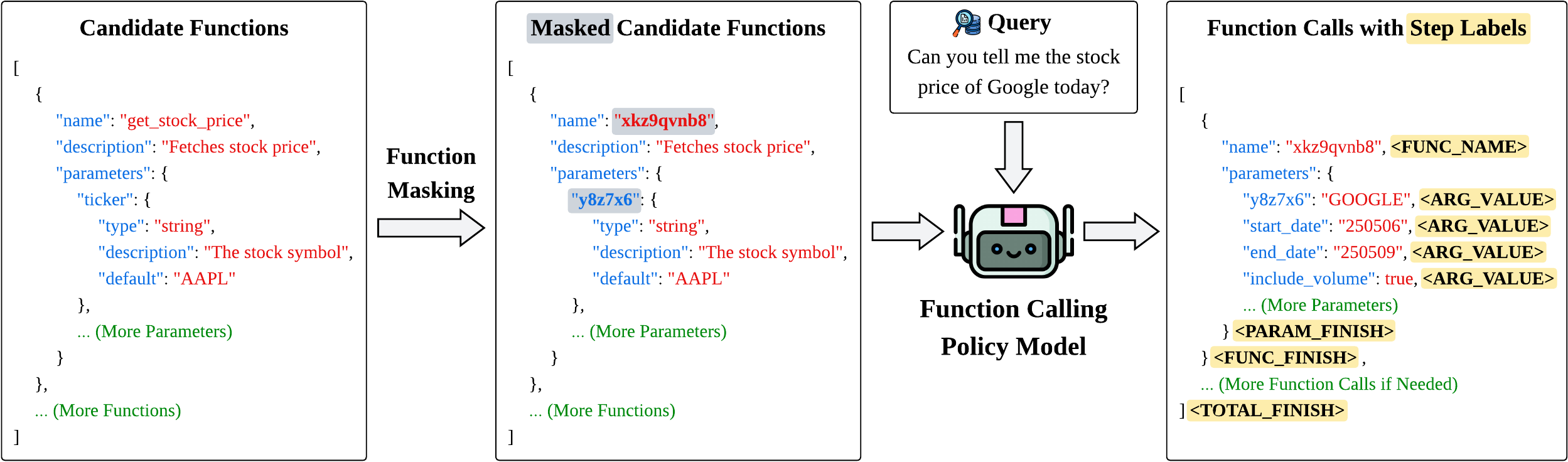}
    \caption{The illustration of data collection for ToolPRM.}
    \label{fig:data collection}
\end{figure*}

\subsection{Fine-Grained Decomposition of Function Calls}

The essence of ToolPRM lies in its fine-grained decomposition of the function calling process. Traditional process supervision approaches for tool invocations primarily rely on coarse-grained response-level rewards, which evaluate the function call as a monolithic unit. 
However, such coarse granularity limits the interpretability, debuggability, and optimization potential.

To this end, ToolPRM decomposes each function call into fine-grained but semantically meaningful steps. Each LLM-generated response comprises a sequence of function calls tailored to a given user query. The construction of each function call is further broken down into (a) selecting the appropriate function name and (b) iteratively identifying parameter names and assigning corresponding values. 
Rather than evaluating a function call as a whole, ToolPRM provides fine-grained supervision that assesses the correctness of each constituent decision. As shown in Figure~\ref{fig:data collection}, this decomposition serves as the foundation for data collection and annotation.

\subsection{Data Collection and Annotation}
\label{sec:data collection annotation}

To enable robust reward modeling under our fine-grained process supervision paradigm, we construct a high-quality annotated dataset that reflects the intricacies of individual decision points.

\paragraph{Data Collection.}
We begin by collecting natural language queries paired with their corresponding structured function calls, using 
xlam-function-calling-60k~\cite{liu2024apigen} and xlam-irrelevance-7.5k~\cite{lin2024hammerrobustfunctioncallingondevice} datasets as the foundations. 
As shown in Figure~\ref{fig:data collection}, we apply function masking that selectively replaces function names and parameter identifiers with random strings. 
The masking mechanism introduces function ambiguity, encouraging the model to rely on contextual understanding of descriptions instead of simple memorization of the tool names, thus enhancing model robustness and generalization~\cite{lin2024hammerrobustfunctioncallingondevice}.
We adopt Hammer2.1-3b and Hammer2.1-7b as policy models to perform the rollout for data collection.
Given a user query and a set of masked function candidates, each rollout generates a sequence of function calls, which is subsequently annotated with fine-grained step labels.

\paragraph{Result Annotation.}
As shown in Figure~\ref{fig:data collection}, the structured function call generated by the policy model is usually in JSON format. 
We can provide fine-grained step labels with the following types:
\begin{itemize}[leftmargin=10pt]
    \item \texttt{<FUNC\_NAME>}: Whether the selected function name at this round is correct or not. 
    \item \texttt{<ARG\_VALUE>}: Whether the one pair of parameter name and value is correctly filled in, which can repeat multiple times.
    \item \texttt{<PARAM\_FINISH>}: Whether all the parameters  and values are correctly assigned.
    \item \texttt{<FUNC\_FINISH>}: Whether the single function call (one element of the list) is correct or not.
    \item \texttt{<TOTAL\_FINISH>}: Whether the overall response (a list of function calls) is correct or not.
\end{itemize}
Each symbol is followed by a binary label annotated by exact match with any of the possible ground truths, to indicate the correctness of the corresponding fine-grained step. 
Some of the step labels seem redundant (\eg, \texttt{<ARG\_VALUE>} and \texttt{<PARAM\_FINISH>}), but our experiments show that such a hierarchical step label redundancy can help the reward model generalize and perform better.

\subsection{State Transition Mechanism and ToolPRM Training}

\begin{figure*}[t]
    \centering
    \includegraphics[width=0.9\textwidth]{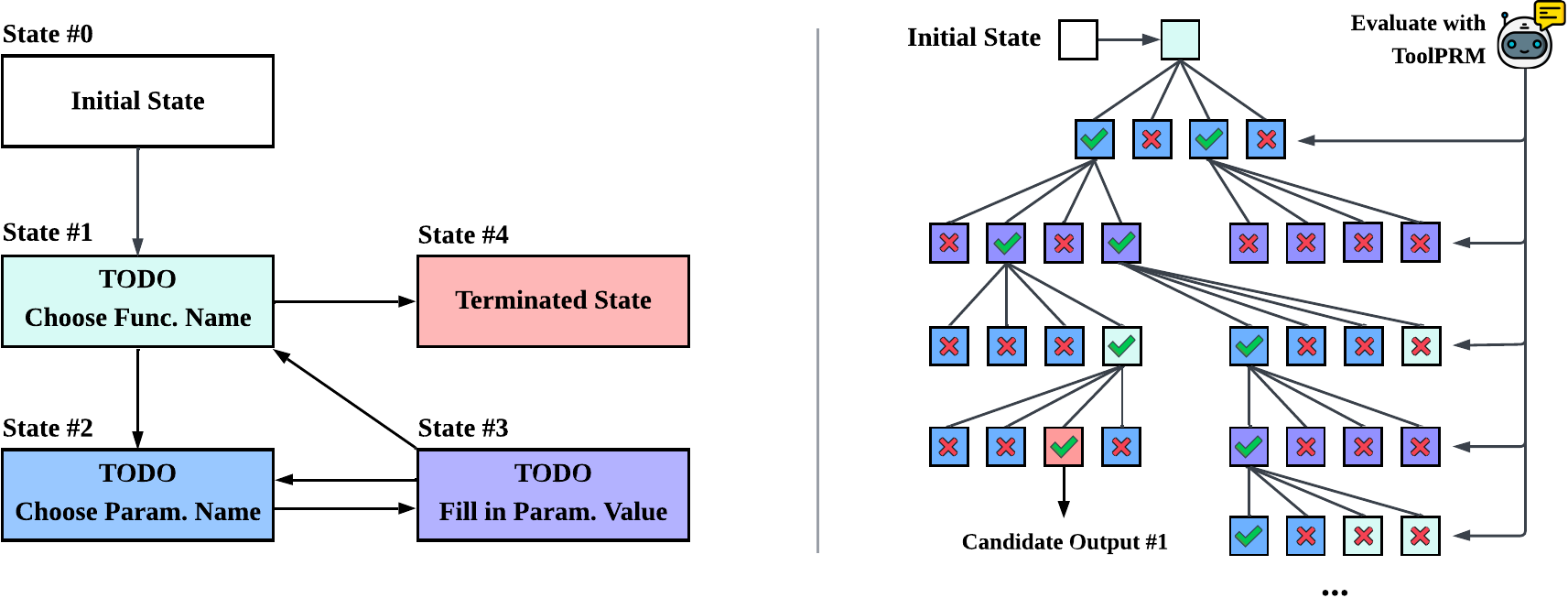}
    \caption{The state transition of function calling (left) and beam search with ToolPRM (right).}
    \label{fig:method}
\end{figure*}

Based on the fine-grained step labels above, we can further formalize the function calling generation process as a dynamic decision process with a series of state transitions. 
As illustrated in Figure~\ref{fig:method}, we define the following five fine-grained states:
\begin{itemize}[leftmargin=10pt]
    \item \textbf{State \#0 (Initial State)}: Input context and masked function candidates are presented.
    \item \textbf{State \#1 (To Select Function Name)}: The model is required to select the function name from the candidates that are aligned with the query intent.
    \item \textbf{State \#2 (To Select Parameter Name)}: The model should choose one necessary parameter to be filled in for the selected function.
    \item \textbf{State \#3 (To Fill in Parameter Value)}: The model has to assign the proper value to the previously selected function parameter.
    \item \textbf{State \#4 (Terminated State)}: One LLM response for function calls completes.
\end{itemize}
Each transition between states can be explicitly supervised by the fine-grained step labels discussed in Section "Data Collection and Annotation for Reward Modeling".

To train ToolPRM under fine-grained process supervision, we represent the function calling generation as a trajectory of decision steps, each labeled with a binary process reward.

Let $\mathcal{T} = \{(s_t, a_t, r_t)\}_{t=1}^{T}$ denote a trajectory, where $s_t$ is the state at step $t$, $a_t$ is the action (\eg, selecting a function or filling a parameter via language token generation), and $r_t \in \{0, 1\}$ indicates its correctness. 
Each state encodes the current decision context, including the input query, masked function candidates, as well as the partially generated function calls. 

In this paper, since the backbone of ToolPRM is also a large language model, we adopt the tokens ``+'' and ``-'' for positive and negative reward labels, respectively. 
Given $N$ annotated trajectories $\{\mathcal{T}_i\}_{i=1}^{N}$, ToolPRM is trained to predict $r_t$ for each $(s_t, a_t)$ via generative process reward modeling:
\begin{equation}
    \mathcal{L}_{\mathrm{ToolPRM}} = - \mathbb{E}_{\tau\in\mathcal{D},(s_t,a_t,r_t)\in\tau}
    \log p_{\theta}(r_t|s_t,a_t),
\end{equation}
where $r_t^{(i)} \in \{+,-\}$ is the binary label token, and $\theta$ is the parameters of backbone LLM.

\subsection{Beam Search with Fine-Grained Process Supervision}
\label{sec: process beam search}

As shown in Figure~\ref{fig:method}, we apply beam search guided by ToolPRM to generate high-quality structured outputs. At each step, ToolPRM assigns a fine-grained process reward to candidate actions based on their local state-action pair, allowing us to prune incorrect partial trajectories early.
Specifically, we compute the score $s$ for each beam candidate with ToolPRM \ie, $s=e^{s_{+}}/(e^{s_{+}}+e^{s_{-}})$, where $s_{+}$ and $s_{-}$ are the predictive logits on label tokens $\{+,-\}$.
We maintain the top-$N$ highest-scoring candidates, where $N$ indicates the number of beams. 
At each step, every preserved candidate can generate $M$ subsequent steps for ToolPRM evaluation, where $M$ denotes the beam width. 

We propose a core principle for inference scaling in structured output generation: \textbf{explore more but retain less}. Concretely, we increase beam width $M$ to explore a wider range of candidate trajectories, but retain only a small number $N$ of highly promising ones. This design reflects the unrecoverability of structured outputs, such as function calls in JSON format, where a single incorrect decision (e.g., a wrong function name or argument value) can invalidate the entire trajectory.

This stands in contrast to inference scaling in unstructured tasks such as math reasoning~\cite{snell2024scaling} or free-form text generation~\cite{setlur2024rewarding,zhang2025igniting}, where early errors can often be corrected or compensated for in later steps. In such settings, preserving a diverse set of candidates throughout decoding (i.e., larger $N$) is beneficial. However, in structured generation, most decision points admit only a single or very few valid actions. Retaining incorrect candidates leads to inefficient use of the generation budget, as future steps cannot recover from early structural mistakes.

Therefore, expanding the search space (larger beam width $M$) while aggressively pruning based on ToolPRM's step-wise supervision (smaller number of beams $N$) ensures that computational resources are concentrated on valid, high-quality structured outputs. This principle underpins the effectiveness of ToolPRM in scaling inference for structured generation tasks.
\section{Experiments}

\subsection{Experiment Setups}

\begin{table}[t]
\centering
\small
\caption{The statistics of our ToolPRM dataset.}
\begin{tabular}{cccc}
\toprule
\multirow{2}{*}{\begin{tabular}[c] {@{}c@{}} Sample Granularity \\ (Data Split)\end{tabular}}
 & \multirow{2}{*}{\begin{tabular}[c] {@{}c@{}} Positive \end{tabular}} & \multirow{2}{*}{\begin{tabular}[c] {@{}c@{}} Negative \end{tabular}} & \multirow{2}{*}{\begin{tabular}[c] {@{}c@{}} Total \end{tabular}} \\ 
 & & & \\\midrule
Step (Train) & 4,380,323 & 731,665 & 5,111,988\\
Trajectory (Train) & 466,786 & 127,648 & 594,434\\
Step (Test) & 488,611 & 81,366 & 569,977 \\
Trajectory (Test) & 52,030 & 14,019 & 66,049 \\
\bottomrule
\end{tabular}
\label{tab:dataset_statistics}
\end{table}

\paragraph{Datasets.}

To evaluate our proposed ToolPRM for function calling, we process and annotate the xlam-function-calling-60k~\cite{liu2024apigen} and xlam-irrelevance-7.5k~\cite{lin2024hammerrobustfunctioncallingondevice} datasets using a well designed pipeline, as detailed in Section ``Methodology''. 
We report the dataset statistics in Table~\ref{tab:dataset_statistics}. 
The resulting annotated dataset exhibits an average of 6.25 step labels per sample, culminating in a total of 192,061 samples. 
The integration of  function masking techniques further expands the dataset size to 4 to 5 times, bringing it to a scale that surpasses typical datasets for Math PRM development, such as OpenAI's publicly available prm800k dataset~\cite{lightman2023letsverifystepstep}.

Note that our constructed dataset above is used to train and validate the predictive accuracy of reward models. 
To further validate the effectiveness of our inference scaling strategy using ToolPRM, we adopt two function-calling benchmarks: BFCL (Berkeley Function Calling Leaderboard)~\cite{berkeley-function-calling-leaderboard} and ToolAlpaca~\cite{tang2023toolalpaca}. 
Following previous works~\cite{abdelaziz2024granite,lin2024hammerrobustfunctioncallingondevice}, we use abstract-syntax-tree-based (AST-based) accuracy as the metric for BFCL, and employ F1 scores of both API selection and parameter value assignment for ToolAlpaca.

\paragraph{Baselines.}
We compare our proposed ToolPRM to different sets of baselines from two aspects. 

To evaluate the predictive accuracy improvement brought by fine-grained intra-call process supervision, we train two distinct baseline models: outcome reward model (ORM) and coarse-grained process reward model (C-PRM).
Compared with our ToolPRM, \ie, fine-grained process reward model using all the step labels, these two baselines are trained on the same dataset with different choices of step labels:
\begin{itemize}[leftmargin=10pt]
    \item \textbf{ORM} is trained to discriminate the final result of the entire function-calling sequence with \texttt{<TOTAL\_FINISH>}. All the other intermediate step labels are disregarded.
    \item \textbf{C-PRM} offers a more granular assessment of function call correctness than ORM by considering all the step labels except \texttt{<ARG\_VALUE>}.
\end{itemize}

To evaluate the function-calling capabilities, we adopt the following three types of baselines:
\begin{itemize}[leftmargin=10pt]
    \item \textbf{General Purpose Models.} This group comprises large language models that have not undergone specific finetuning for function calling tasks, nor do they employ the inference scaling strategies. 
    Their performance serves as a baseline representing general capabilities. 
    The models include GPT-4o, GPT-4o-mini \cite{openai2024gpt4ocard}, Llama-3.1-8B-Instruct \cite{grattafiori2024llama3herdmodels}, Mistral-Nemo-Instruct \cite{jiang2023mistral7b}, and the Qwen2.5-Instruct series (72B, 32B, 7B, 3B, 1.5B) \cite{qwen2.5}. 
    
    \item \textbf{Function Calling Models}. This category includes models that have been specifically designed or finetuned for function calling tasks. 
    These models include GRANITE (GRANITE-20B-FUNCTIONCALLING)~\cite{abdelaziz2024granite}, xLAM-fc-r series (7B, 1B)~\cite{liu2024apigen}, OpenFunctions-v2~\cite{patil2023gorillalargelanguagemodel}, and Hammer2.1 series (7B, 3B, 1.5B, 0.5B)~\cite{lin2024hammerrobustfunctioncallingondevice}.

    \item \textbf{Inference Scaling Strategies.} We evaluate the performance enhancements afforded by different inference scaling methods. We choose the Hammer2.1 series (7B, 3B, and 1.5B variants) as base policy models. 
    Three distinct inference scaling strategies are applied and compared against our ToolPRM: token-level beam search, majority voting, and best of N.
\end{itemize}


\paragraph{Implementation Details.} 

All the experiments are conducted on NVIDIA 8xH100 GPU Clusters.
We use Hammer2.1-3b as the reward model backbone and adopt SFT to train ToolPRM for 5 epochs with Adam optimizer. 
The batch size is 1024. 
The learning rate is 1e-3 with a warmup ratio of 0.008 followed by linear learning rate decay.
The weight decay is 1e-5.
As for the fine-grained beam search with ToolPRM, we set the temperature as 0.8, and select the number of beams $N$ and the beam width $M$ from $\{1, 2, 4, 8, 16\}$.


\subsection{Predictive Accuracy of Reward Model}

High predictive accuracy of the reward model is vital for increasing the probability of selecting and retaining correct function call sequences during inference scaling, ultimately leading to superior overall inference outcomes.
Hence, this section investigates the performance of reward models for function calls with varying granularities. 

We use three metrics for this evaluation: model loss, step-level accuracy (Step Acc), and trajectory-level accuracy (Trajectory Acc). 
Step Acc quantifies the correctness at each discrete process supervision step. 
It is worth noting that since each reward model (ORM, C-PRM, and ToolPRM) is supervised at its distinct level of granularity, the inherent difficulty of accurately predicting each step differs. 
Hence, we introduce trajectory-level accuracy to facilitate an apple-to-apple comparison of models trained with differing supervisory step labels. 
The trajectory accuracy evaluates the correctness of the entire function call sequence. A trajectory is deemed accurate if the final assessment of the entire sequence is correct. This means that for C-PRM and ToolPRM, even if their evaluations of some intermediate steps are incorrect, the trajectory can still be classified accurately if their overall judgment of the function call's success or failure is correct.

\begin{table}[t]
\centering
\small
\caption{Predicting accuracy of RMs of different granularities}
\begin{tabular}{ccccc}
\toprule
Reward Model &  Loss & Step Acc & Trajectory Acc \\
\midrule
ORM  & 0.0536 & 98.39\% & 98.39\%\\
C-PRM & 0.0371 & 98.87\% & 99.06\% \\
ToolPRM & \textbf{0.0286} & \textbf{99.11\%} & \textbf{99.38\%} \\
\bottomrule
\end{tabular}
\label{tab:granularity_comparison}
\end{table}
\begin{figure}[t] 
    \centering 
    \includegraphics[width=0.46\textwidth]{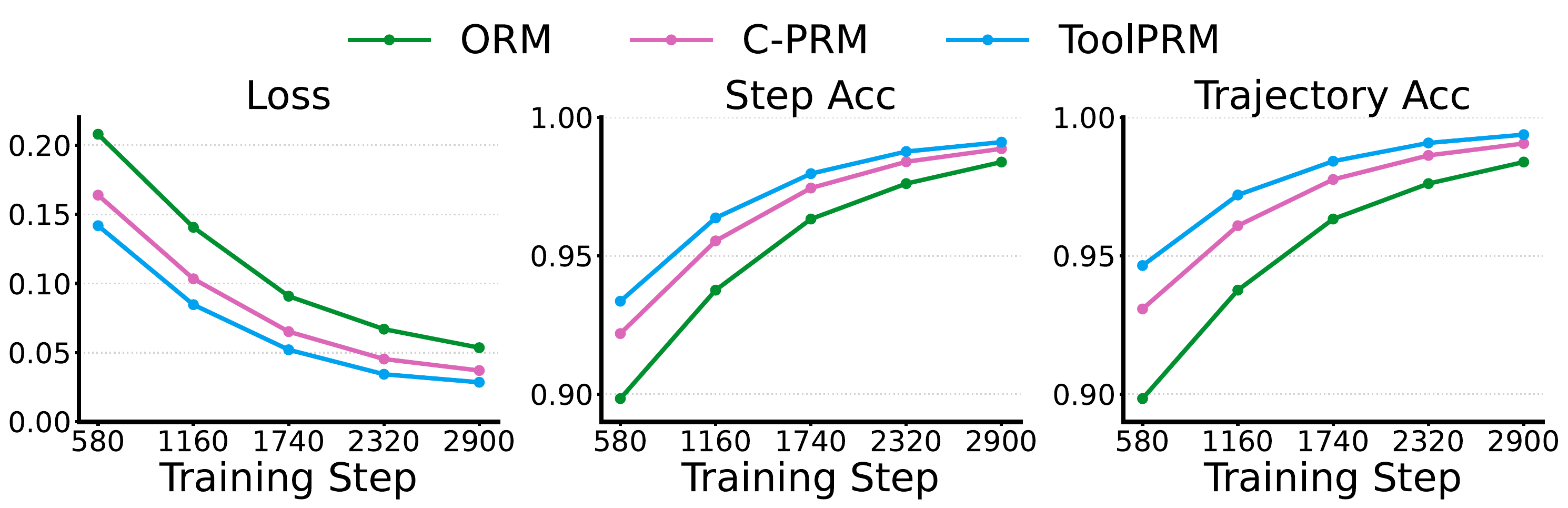}
    \caption{The 5-epoch learning curves of different RMs (\ie, ORM, C-PRM, ToolPRM) in terms of loss, step-level accuracy, and trajectory-level accuracy.}
    \vspace{-10pt}
    \label{fig:rm learning curve} 
\end{figure}

We report the predictive accuracy results in Table~\ref{tab:granularity_comparison} and illustrate the learning curves over the five-epoch training in Figure~\ref{fig:compare_PRMs}. 
We can observe that reward models with finer granularity consistently achieve higher predictive accuracy across all three metrics. 
Specifically, our proposed ToolPRM outperforms both C-PRM and ORM in this regard. 
This suggests that fine-grained process reward modeling, such as ToolPRM, offers superior performance not only in terms of the precision of step-level supervision but also in their ultimate effectiveness in judging overall sequence correctness compared to their coarser-grained counterparts.

\subsection{Inference Scaling Performance}






We conduct a comprehensive examination and comparative analysis of our ToolPRM, as well as various function calling baseline models and inference scaling strategies.

As evidenced in Table \ref{tab:main_result}, we can observe that our proposed ToolPRM can generally achieve the best performance compared with other inference scaling strategies given base policy models of different sizes. 
Besides, the performance of other inference scaling strategies is fairly unstable and can sometimes perform worse than the base model.
The potential reason is that the non-greedy sampling might cause minor errors during the function calling generation, which could directly ruin the entire trajectory. 
This suggests the necessity of intra-call process supervision through the fine-grained beam search for structured function calling generation.

Notably, the performance uplift brought by our ToolPRM is more pronounced for smaller policy models. This characteristic renders ToolPRM particularly advantageous for on-device inference scenarios, which is of considerable importance for applications such as function calling that are frequently deployed in edge environments and are critical for real-world utility.

Illustratively, the Hammer2.1-1.5B model, when augmented with our ToolPRM method, achieves performance comparable to the baseline 3B model. Similarly, applying ToolPRM to the Hammer2.1-3B model elevates its performance to a level on par with the baseline 7B model. Furthermore, the integration of ToolPRM with the Hammer2.1-7B model enables it to outperform several significantly larger models that previously held a performance advantage, including state-of-the-art models such as Qwen2.5-32B-Instruct.

These empirical results validate the effectiveness of our proposed inference scaling framework for structured function calling generation. Applying fine-grained beam search with ToolPRM can substantially improve the capability of base models at inference time through computational scaling.


\begin{table*}[t]  \small
\centering
\caption{Performance comparison of the general purpose models, function calling models, and different inference scaling techniques applied on Hammer2.1 series of models, evaluated on BFCL and ToolAlpaca. Multi., Paral., Mul.P. represents the Multiple split, Parallel split, and Multiple Parallel split of BFCL, separately. And Avg. represents an unweighted average of all the sub-categories of that benchmark. The best results of each model type (\ie, general purpose, function calling, and inference scaling) is given in bold and the second best is underlined.}
\resizebox{0.96\textwidth}{!}{
\renewcommand\arraystretch{1.03}
\begin{tabular}{clccccccccc}
\toprule
\multirow{2}{*}{Type} & \multirow{2}{*}{Model} & \multirow{2}{*}{Size} & \multicolumn{5}{c}{BFCL} & \multicolumn{3}{c}{ToolAlpaca} \\ \cmidrule(lr){4-8} \cmidrule(lr){9-11} 
 &  &  & Simple & Multi. & Paral. & Mul.P. & Avg. & F1-API & F1-Args & Avg. \\ \midrule
\multirow{9}{*}{\begin{tabular}[c]{@{}c@{}}General\\      Purpose\end{tabular}} & GPT-4o & - & 77.17 & \underline{95.00} & \textbf{93.50} & 85.00 & \underline{87.67} & \textbf{88.64} & \textbf{66.67} & \textbf{77.66} \\
 & GPT-4o-mini & - & \underline{80.08} & 90.50 & 89.50 & 87.00 & 86.77 & 64.34 & 54.69 & 59.52 \\
 & Llama-3.1-8B-Instruct & 8B & 72.83 & 93.50 & 87.00 & 83.50 & 84.21 & 75.64 & 55.12 & 65.38 \\
 & Mistral-Nemo-Instruct & 12B & 77.00 & 93.50 & 89.50 & 84.50 & 86.13 & \underline{87.31} & \underline{66.18} & \underline{76.75} \\
 & Qwen2.5-72B-Instruct & 72B & \textbf{80.25} & \textbf{97.50} & \textbf{93.50} & \textbf{92.00} & \textbf{90.81} & 83.21 & 65.56 & 74.39 \\
 & Qwen2.5-32B-Instruct & 32B & 72.83 & 94.00 & \textbf{93.50} & \underline{88.50} & 87.21 & 85.82 & 61.11 & 73.47 \\
 & Qwen2.5-7B-Instruct & 7B & 75.33 & 94.50 & \underline{91.50} & 84.50 & 86.46 & 83.70 & 60.00 & 71.85 \\
 & Qwen2.5-3B-Instruct & 3B & 74.17 & 90.50 & 79.50 & 79.00 & 80.79 & 70.80 & 51.63 & 61.22 \\
 & Qwen2.5-1.5B-Instruct & 1.5B & 72.42 & 87.00 & 81.50 & 75.50 & 79.11 & 62.07 & 43.23 & 52.65 \\ \midrule
\multirow{8}{*}{\begin{tabular}[c]{@{}c@{}}Function\\      Calling\end{tabular}} & GRANITE & 20B & 72.83 & 91.50 & 84.00 & 81.50 & 82.46 & 77.27 & 58.00 & 67.64 \\
 & xLAM-7b-fc-r & 7B & 73.08 & \underline{93.50} & 87.00 & \underline{84.00} & 84.40 & 67.26 & 58.96 & 63.11 \\
 & xLAM-1b-fc-r & 1.3B & 69.67 & 89.50 & 79.00 & 66.50 & 76.17 & 64.86 & 50.58 & 57.72 \\
 & OpenFunctions-v2 & 7B & \textbf{83.27} & 93.00 & 85.50 & 66.00 & 81.94 & 72.93 & 51.26 & 62.10 \\
 & Hammer2.1-7B & 7B & 78.08 & \textbf{95.00} & \textbf{93.50} & \textbf{88.00} & \textbf{88.65} & \textbf{80.93} & \textbf{64.60} & \textbf{72.77} \\
 & Hammer2.1-3B & 3B & \underline{81.42} & \textbf{95.00} & \underline{89.50} & 81.50 & \underline{86.86} & \underline{80.31} & \underline{62.83} & \underline{71.57} \\
 & Hammer2.1-1.5B & 1.5B & 74.67 & 92.00 & 84.50 & 80.00 & 82.79 & 77.42 & 61.17 & 69.30 \\
 & Hammer2.1-0.5B & 0.5B & 68.00 & 83.00 & 71.50 & 54.00 & 69.13 & 77.10 & 60.67 & 68.89 \\ \midrule
\multirow{15}{*}{\begin{tabular}[c]{@{}c@{}}Inference\\      Scaling\end{tabular}} & Hammer2.1-7B (Base) & \multirow{5}{*}{7B} & 78.08 & \underline{95.00} & \underline{93.50} & \underline{88.00} & \underline{88.65} & \underline{80.93} & 64.60 & 72.77 \\
 &+ Token-level Beam Search &  & 74.58 & 93.50 & 91.50 & 82.50 & 85.52 & 79.69 & 62.37 & 71.03 \\
 &+ Marjority &  & \textbf{79.58} & \underline{95.00} & \underline{93.50} & 85.00 & 88.27 & 79.03 & 65.26 & 72.15 \\
 &+ Best of N (ORM) &  & 76.58 & 94.50 & 91.50 & 87.00 & 87.40 & 78.86 & \textbf{67.73} & \underline{73.30} \\
 &\cellcolor{gray!20}+ ToolPRM  (Ours) &  & \cellcolor{gray!20}\underline{79.08} & \cellcolor{gray!20}\textbf{95.50} & \cellcolor{gray!20}\textbf{94.50} & \cellcolor{gray!20}\textbf{89.00} & \cellcolor{gray!20}\textbf{89.52} & \cellcolor{gray!20}\textbf{81.42} & \cellcolor{gray!20}\underline{65.30} & \cellcolor{gray!20}\textbf{73.36} \\
 \addlinespace[-0.5 \aboverulesep]
 \cmidrule{2-11}
 \addlinespace[-0.5 \belowrulesep]
 & Hammer2.1-3B (Base) & \multirow{5}{*}{3B} & \textbf{81.42} & \underline{95.00} & \underline{89.50} & 81.50 & \underline{86.86} & \underline{80.31} & \underline{62.83} & \underline{71.57} \\
 &+ Token-level Beam Search &  & 74.50 & 91.00 & 87.50 & 77.00 & 82.50 & 78.29 & 59.32 & 68.81 \\
 &+ Marjority &  & 79.50 & \underline{95.00} & 89.00 & 81.00 & 86.13 & 76.00 & 58.27 & 67.14 \\
 &+ Best of N (ORM) &  & 77.08 & 93.50 & \underline{89.50} & \underline{84.00} & 86.02 & 77.29 & 60.16 & 68.73 \\
 &\cellcolor{gray!20}+ ToolPRM   (Ours) &  & \cellcolor{gray!20}\underline{80.50} & \cellcolor{gray!20}\textbf{95.50} & \cellcolor{gray!20}\textbf{91.50} & \cellcolor{gray!20}\textbf{88.00} & \cellcolor{gray!20}\textbf{88.88} & \cellcolor{gray!20}\textbf{80.78} & \cellcolor{gray!20}\textbf{63.13} & \cellcolor{gray!20}\textbf{71.96} \\
 \addlinespace[-0.5 \aboverulesep]
 \cmidrule{2-11}
 \addlinespace[-0.5 \belowrulesep]
 & Hammer2.1-1.5B (Base) & \multirow{5}{*}{1.5B} & 74.67 & \textbf{92.00} & 84.50 & 80.00 & 82.79 & \underline{77.42} & 61.17 & 69.30 \\
 &+ Token-level Beam Search &  & 72.33 & \underline{91.50} & 81.00 & 73.50 & 79.58 & 73.03 & 56.68 & 64.86 \\
 &+ Marjority &  & \underline{77.50} & \underline{91.50} & 84.50 & 80.00 & 83.38 & 72.88 & \textbf{63.61} & 68.25 \\
 &+ Best of N (ORM) &  & 75.25 & 90.00 & \underline{86.50} & \underline{84.00} & \underline{83.94} & \underline{77.42} & 61.66 & \underline{69.54} \\
 &\cellcolor{gray!20}+ ToolPRM   (Ours) &  & \cellcolor{gray!20}\textbf{78.42} & \cellcolor{gray!20}\textbf{92.00} & \cellcolor{gray!20}\textbf{87.50} & \cellcolor{gray!20}\textbf{84.50} & \cellcolor{gray!20}\textbf{85.61} & \cellcolor{gray!20}\textbf{82.68} & \cellcolor{gray!20}\underline{63.18} & \cellcolor{gray!20}\textbf{72.93} \\
 \bottomrule
\end{tabular}
}
\label{tab:main_result}
\end{table*}

\subsection{In-Depth Analysis}

\begin{figure}[t] 
    \centering 
    \includegraphics[width=0.46\textwidth]{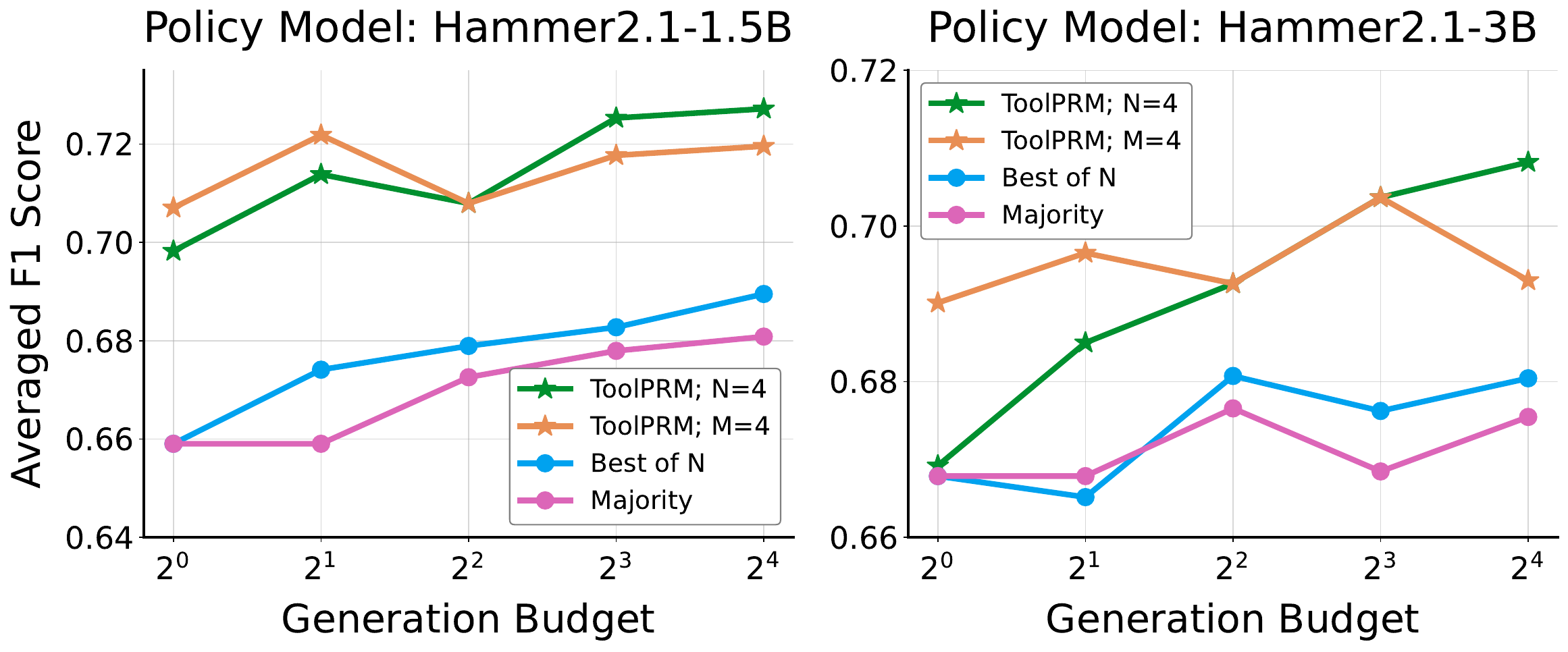}
    \caption{The F1 performance on ToolAlpaca w.r.t. different generation budgets and inference scaling strategies. We conduct experiments on Hammer2.1-1.5B (left) and Hammer2.1-3B (right) policy models.
    }
    \vspace{-10pt}
    \label{fig:compare_PRMs} 
\end{figure}

In this section, we investigate the performance increase brought by the scaling of generation budget. We choose Hammer2.1-1.5B and Hammer2.1-3B as the base policy models, and conduct experiments on ToolAlpaca dataset. 
We report the averaged F1 score over both function selection and parameter value assignment in Figure \ref{fig:compare_PRMs}.

For majority voting and best of N approaches, the generation budget refers to the number of candidate trajectories to be sampled. 
For our ToolPRM, as discussed in Section ``Beam Search with Fine-Grained Process
Supervision via ToolPRM'', there are two key hyperparameters that can determine the generation budget, \ie, the number of beams $N$ and the beam width $M$.
Hence, we develop two ToolPRM variants by fixing the one hyperparameter as 4 and increasing the other for a larger budget.

We first compare ToolPRM variants with $N=4$ and $M=4$. 
When fixing the number of candidates to be retrained at 4 and scaling the beam width, it generally demonstrates a consistent improvement in accuracy as the generation budget increases. 
Conversely, when maintaining a fixed beam width of 4 and scaling the number of candidates, less performance gain is exhibited, and in some instances, increased N even leads to a noticeable degradation in accuracy. 
This phenomenon validates our core insights that increased retention of wrong intermediate steps (\ie, larger $N$) can misguide subsequent generation steps. 
With higher generation budgets, the benefits derived from expanded exploration (scaling beam width) tend to outweigh those derived from increased retention (scaling N), \ie, explore more but retain less.

Moreover, our ToolPRM approaches generally demonstrate superior performance when compared to the baseline strategies ``Best of N'' and ``Majority Voting''. 
The ``Best of N'' strategy shows a modest increase in averaged F1 score with a larger budget, while the 'Majority' strategy generally lags behind. 
Overall, fine-grained beam search with ToolPRM appears to be a robust approach, particularly when a larger generation budget is available and the ``explore more but retain less'' principle is followed, highlighting the importance of exploration guided by a fine-grained process reward model for structured function calling generation.
\section{Conclusion}

In this paper, we propose a fine-grained inference scaling framework to enhance LLM performance on structured function calling tasks. We construct an intra-call step-annotated dataset to train ToolPRM, a process reward model that supervises each intermediate step. 
Integrated with beam search, ToolPRM achieves the highest supervision accuracy and enables base LLMs to attain state-of-the-art results. 
We also identify a key principle for structured inference scaling: ``explore more but retain less'' based on the unrecoverability of structured output generation. 
However, the optimal trade-off in this principle is not yet dynamically adjustable. Future work could explore adaptive strategies that calibrate exploration and retention based on input complexity or ToolPRM-derived confidence. 
\section*{Limitations}
While ToolPRM improves fine-grained guidance for tool use, it still assumes a discretized, step-wise view of decision making, which may not capture all forms of implicit reasoning or latent uncertainty. The approach focuses on rewarding intermediate structure and consistency, and thus cannot guarantee global optimality of the final tool choice or argument specification in every case. In addition, the framework introduces additional modeling components (e.g., masking design and state definitions), whose choices may affect behavior and require careful implementation. 

\section*{Acknowledgements}

We would like to thank the anonymous reviewers for their valuable comments and suggestions.
This work was supported in part by the National Key R\&D Program of China (2024YFC3505402) and the National Natural Science Foundation of China (624B2096,62322603,U2244217,72542012,72595872).

\bibliography{custom}

@article{wu2025inferencescalinglawsempirical,
  title={Inference scaling laws: An empirical analysis of compute-optimal inference for problem-solving with language models},
  author={Wu, Yangzhen and Sun, Zhiqing and Li, Shanda and Welleck, Sean and Yang, Yiming},
  journal={arXiv preprint arXiv:2408.00724},
  year={2024}
}

@article{ouyang2022traininglanguagemodelsfollow,
  title={Training language models to follow instructions with human feedback},
  author={Ouyang, Long and Wu, Jeffrey and Jiang, Xu and Almeida, Diogo and Wainwright, Carroll and Mishkin, Pamela and Zhang, Chong and Agarwal, Sandhini and Slama, Katarina and Ray, Alex and others},
  journal={Advances in neural information processing systems},
  volume={35},
  pages={27730--27744},
  year={2022}
}

@article{chai2025parl,
  title={PARL-MT: Learning to Call Functions in Multi-Turn Conversation with Progress Awareness},
  author={Chai, Huacan and Cao, Zijie and Ran, Maolin and Yang, Yingxuan and Lin, Jianghao and Peng, Xin and Wang, Hairui and Ding, Renjie and Wan, Ziyu and Wen, Muning and others},
  journal={arXiv preprint arXiv:2509.23206},
  year={2025}
}

@article{zheng2025survey,
  title={A survey of process reward models: From outcome signals to process supervisions for large language models},
  author={Zheng, Congming and Zhu, Jiachen and Ou, Zhuoying and Chen, Yuxiang and Zhang, Kangning and Shan, Rong and Zheng, Zeyu and Yang, Mengyue and Lin, Jianghao and Yu, Yong and others},
  journal={arXiv preprint arXiv:2510.08049},
  year={2025}
}

@article{zheng2025cold,
  title={Cold: Counterfactually-guided length debiasing for process reward models},
  author={Zheng, Congmin and Zhu, Jiachen and Lin, Jianghao and Dai, Xinyi and Yu, Yong and Zhang, Weinan and Yang, Mengyue},
  journal={arXiv preprint arXiv:2507.15698},
  year={2025}
}

@inproceedings{zhu2025retrieval,
  title={Retrieval-Augmented Process Reward Model for Generalizable Mathematical Reasoning},
  author={Zhu, Jiachen and Zheng, Congmin and Lin, Jianghao and Du, Kounianhua and Wen, Ying and Yu, Yong and Wang, Jun and Zhang, Weinan},
  booktitle={Findings of the Association for Computational Linguistics: ACL 2025},
  pages={8453--8468},
  year={2025}
}

@article{xi2025survey,
  title={A survey of llm-based deep search agents: Paradigm, optimization, evaluation, and challenges},
  author={Xi, Yunjia and Lin, Jianghao and Xiao, Yongzhao and Zhou, Zheli and Shan, Rong and Gao, Te and Zhu, Jiachen and Liu, Weiwen and Yu, Yong and Zhang, Weinan},
  journal={arXiv preprint arXiv:2508.05668},
  year={2025}
}

@article{yang2025survey,
  title={A survey of ai agent protocols},
  author={Yang, Yingxuan and Chai, Huacan and Song, Yuanyi and Qi, Siyuan and Wen, Muning and Li, Ning and Liao, Junwei and Hu, Haoyi and Lin, Jianghao and Chang, Gaowei and others},
  journal={arXiv preprint arXiv:2504.16736},
  year={2025}
}

@article{zhou2025steporlm,
  title={Steporlm: A self-evolving framework with generative process supervision for operations research language models},
  author={Zhou, Chenyu and Xu, Tianyi and Lin, Jianghao and Ge, Dongdong},
  journal={arXiv preprint arXiv:2509.22558},
  year={2025}
}

@article{zhou2026externalization,
  title={Externalization in LLM Agents: A Unified Review of Memory, Skills, Protocols and Harness Engineering},
  author={Zhou, Chenyu and Chai, Huacan and Chen, Wenteng and Guo, Zihan and Shan, Rong and Song, Yuanyi and Xu, Tianyi and Yang, Yingxuan and Yu, Aofan and Zhang, Weiming and others},
  journal={arXiv preprint arXiv:2604.08224},
  year={2026}
}

@article{uesato2022solvingmathwordproblems,
  title={Solving math word problems with process-and outcome-based feedback},
  author={Uesato, Jonathan and Kushman, Nate and Kumar, Ramana and Song, Francis and Siegel, Noah and Wang, Lisa and Creswell, Antonia and Irving, Geoffrey and Higgins, Irina},
  journal={arXiv preprint arXiv:2211.14275},
  year={2022}
}

@inproceedings{lightman2023letsverifystepstep,
  title={Let's verify step by step},
  author={Lightman, Hunter and Kosaraju, Vineet and Burda, Yuri and Edwards, Harrison and Baker, Bowen and Lee, Teddy and Leike, Jan and Schulman, John and Sutskever, Ilya and Cobbe, Karl},
  booktitle={The Twelfth International Conference on Learning Representations},
  year={2023}
}

@inproceedings{berkeley-function-calling-leaderboard,
    title={Berkeley Function Calling Leaderboard},
    author={Fanjia Yan and Huanzhi Mao and Charlie Cheng-Jie Ji and Tianjun Zhang and Shishir G. Patil and Ion Stoica and Joseph E. Gonzalez},
    year={2024},
    howpublished={\url{https://gorilla.cs.berkeley.edu/blogs/8_berkeley_function_calling_leaderboard.html}},
    }

@article{ma2023letsrewardstepstep,
  title={Let's reward step by step: Step-Level reward model as the Navigators for Reasoning},
  author={Ma, Qianli and Zhou, Haotian and Liu, Tingkai and Yuan, Jianbo and Liu, Pengfei and You, Yang and Yang, Hongxia},
  journal={arXiv preprint arXiv:2310.10080},
  year={2023}
}

@article{ke2025surveyfrontiersllmreasoning,
  title={A Survey of Frontiers in LLM Reasoning: Inference Scaling, Learning to Reason, and Agentic Systems},
  author={Ke, Zixuan and Jiao, Fangkai and Ming, Yifei and Nguyen, Xuan-Phi and Xu, Austin and Long, Do Xuan and Li, Minzhi and Qin, Chengwei and Wang, Peifeng and Savarese, Silvio and others},
  journal={arXiv preprint arXiv:2504.09037},
  year={2025}
}

@article{wang2023selfconsistencyimproveschainthought,
  title={Self-consistency improves chain of thought reasoning in language models},
  author={Wang, Xuezhi and Wei, Jason and Schuurmans, Dale and Le, Quoc and Chi, Ed and Narang, Sharan and Chowdhery, Aakanksha and Zhou, Denny},
  journal={arXiv preprint arXiv:2203.11171},
  year={2022}
}

@article{puri2025probabilisticinferenceapproachinferencetime,
  title={A Probabilistic Inference Approach to Inference-Time Scaling of LLMs using Particle-Based Monte Carlo Methods},
  author={Puri, Isha and Sudalairaj, Shivchander and Xu, Guangxuan and Xu, Kai and Srivastava, Akash},
  journal={arXiv preprint arXiv:2502.01618},
  year={2025}
}

@article{brown2024largelanguagemonkeysscaling,
  title={Large language monkeys: Scaling inference compute with repeated sampling},
  author={Brown, Bradley and Juravsky, Jordan and Ehrlich, Ryan and Clark, Ronald and Le, Quoc V and R{\'e}, Christopher and Mirhoseini, Azalia},
  journal={arXiv preprint arXiv:2407.21787},
  year={2024}
}

@article{yao2023treethoughtsdeliberateproblem,
  title={Tree of thoughts: Deliberate problem solving with large language models},
  author={Yao, Shunyu and Yu, Dian and Zhao, Jeffrey and Shafran, Izhak and Griffiths, Tom and Cao, Yuan and Narasimhan, Karthik},
  journal={Advances in neural information processing systems},
  volume={36},
  pages={11809--11822},
  year={2023}
}

@article{zhang2023planninglargelanguagemodels,
  title={Planning with large language models for code generation},
  author={Zhang, Shun and Chen, Zhenfang and Shen, Yikang and Ding, Mingyu and Tenenbaum, Joshua B and Gan, Chuang},
  journal={arXiv preprint arXiv:2303.05510},
  year={2023}
}

@article{zhou2024languageagenttreesearch,
  title={Language agent tree search unifies reasoning acting and planning in language models},
  author={Zhou, Andy and Yan, Kai and Shlapentokh-Rothman, Michal and Wang, Haohan and Wang, Yu-Xiong},
  journal={arXiv preprint arXiv:2310.04406},
  year={2023}
}

@article{ACKLEY1985147,
    title = {A learning algorithm for boltzmann machines},
    journal = {Cognitive Science},
    volume = {9},
    number = {1},
    pages = {147-169},
    year = {1985},
    issn = {0364-0213},
    doi = {https://doi.org/10.1016/S0364-0213(85)80012-4},
    url = {https://www.sciencedirect.com/science/article/pii/S0364021385800124},
    author = {David H. Ackley and Geoffrey E. Hinton and Terrence J. Sejnowski}
}

@article{xie2023selfevaluationguidedbeamsearch,
  title={Self-evaluation guided beam search for reasoning},
  author={Xie, Yuxi and Kawaguchi, Kenji and Zhao, Yiran and Zhao, James Xu and Kan, Min-Yen and He, Junxian and Xie, Michael},
  journal={Advances in Neural Information Processing Systems},
  volume={36},
  pages={41618--41650},
  year={2023}
}

@article{wei2022chain,
  title={Chain-of-thought prompting elicits reasoning in large language models},
  author={Wei, Jason and Wang, Xuezhi and Schuurmans, Dale and Bosma, Maarten and Xia, Fei and Chi, Ed and Le, Quoc V and Zhou, Denny and others},
  journal={Advances in neural information processing systems},
  volume={35},
  pages={24824--24837},
  year={2022}
}

@article{liu2023don,
  title={Don't throw away your value model! Generating more preferable text with Value-Guided Monte-Carlo Tree Search decoding},
  author={Liu, Jiacheng and Cohen, Andrew and Pasunuru, Ramakanth and Choi, Yejin and Hajishirzi, Hannaneh and Celikyilmaz, Asli},
  journal={arXiv preprint arXiv:2309.15028},
  year={2023}
}

@article{choi2023kcts,
  title={KCTS: knowledge-constrained tree search decoding with token-level hallucination detection},
  author={Choi, Sehyun and Fang, Tianqing and Wang, Zhaowei and Song, Yangqiu},
  journal={arXiv preprint arXiv:2310.09044},
  year={2023}
}

@inproceedings{hung2025rewardGuidedTreeSearch,
  title={Reward-Guided Tree Search for Inference Time Alignment of Large Language Models},
  author={Hung, Chia-Yu and Majumder, Navonil and Mehrish, Ambuj and Poria, Soujanya},
  booktitle={Proceedings of the 2025 Conference of the Nations of the Americas Chapter of the Association for Computational Linguistics: Human Language Technologies (Volume 1: Long Papers)},
  pages={12575--12593},
  year={2025}
}

@article{tang2023toolalpaca,
  title={Toolalpaca: Generalized tool learning for language models with 3000 simulated cases},
  author={Tang, Qiaoyu and Deng, Ziliang and Lin, Hongyu and Han, Xianpei and Liang, Qiao and Cao, Boxi and Sun, Le},
  journal={arXiv preprint arXiv:2306.05301},
  year={2023}
}

@article{lin2024hammerrobustfunctioncallingondevice,
  title={Hammer: Robust function-calling for on-device language models via function masking},
  author={Lin, Qiqiang and Wen, Muning and Peng, Qiuying and Nie, Guanyu and Liao, Junwei and Wang, Jun and Mo, Xiaoyun and Zhou, Jiamu and Cheng, Cheng and Zhao, Yin and others},
  journal={arXiv preprint arXiv:2410.04587},
  year={2024}
}

@article{abdelaziz2024granite,
  title={Granite-function calling model: Introducing function calling abilities via multi-task learning of granular tasks},
  author={Abdelaziz, Ibrahim and Basu, Kinjal and Agarwal, Mayank and Kumaravel, Sadhana and Stallone, Matthew and Panda, Rameswar and Rizk, Yara and Bhargav, GP and Crouse, Maxwell and Gunasekara, Chulaka and others},
  journal={arXiv preprint arXiv:2407.00121},
  year={2024}
}

@article{liu2024apigen,
  title={Apigen: Automated pipeline for generating verifiable and diverse function-calling datasets},
  author={Liu, Zuxin and Hoang, Thai and Zhang, Jianguo and Zhu, Ming and Lan, Tian and Tan, Juntao and Yao, Weiran and Liu, Zhiwei and Feng, Yihao and RN, Rithesh and others},
  journal={Advances in Neural Information Processing Systems},
  volume={37},
  pages={54463--54482},
  year={2024}
}

@article{wang2024fantastic,
  title={FANTAstic SEquences and where to find them: Faithful and efficient API call generation through state-tracked constrained decoding and reranking},
  author={Wang, Zhuoer and Ribeiro, Leonardo FR and Papangelis, Alexandros and Mukherjee, Rohan and Wang, Tzu-Yen and Zhao, Xinyan and Biswas, Arijit and Caverlee, James and Metallinou, Angeliki},
  journal={arXiv preprint arXiv:2407.13945},
  year={2024}
}

@article{nath2025toolcomp,
  title={ToolComp: A Multi-Tool Reasoning \& Process Supervision Benchmark},
  author={Nath, Vaskar and Raja, Pranav and Yoon, Claire and Hendryx, Sean},
  journal={arXiv preprint arXiv:2501.01290},
  year={2025}
}

@article{setlur2024rewarding,
  title={Rewarding progress: Scaling automated process verifiers for llm reasoning},
  author={Setlur, Amrith and Nagpal, Chirag and Fisch, Adam and Geng, Xinyang and Eisenstein, Jacob and Agarwal, Rishabh and Agarwal, Alekh and Berant, Jonathan and Kumar, Aviral},
  journal={arXiv preprint arXiv:2410.08146},
  year={2024}
}

@article{zhang2025igniting,
  title={Igniting language intelligence: The hitchhiker’s guide from chain-of-thought reasoning to language agents},
  author={Zhang, Zhuosheng and Yao, Yao and Zhang, Aston and Tang, Xiangru and Ma, Xinbei and He, Zhiwei and Wang, Yiming and Gerstein, Mark and Wang, Rui and Liu, Gongshen and others},
  journal={ACM Computing Surveys},
  volume={57},
  number={8},
  pages={1--39},
  year={2025},
  publisher={ACM New York, NY}
}

@article{snell2024scaling,
  title={Scaling llm test-time compute optimally can be more effective than scaling model parameters},
  author={Snell, Charlie and Lee, Jaehoon and Xu, Kelvin and Kumar, Aviral},
  journal={arXiv preprint arXiv:2408.03314},
  year={2024}
}

@article{liu2024toolace,
  title={Toolace: Winning the points of llm function calling},
  author={Liu, Weiwen and Huang, Xu and Zeng, Xingshan and Hao, Xinlong and Yu, Shuai and Li, Dexun and Wang, Shuai and Gan, Weinan and Liu, Zhengying and Yu, Yuanqing and others},
  journal={arXiv preprint arXiv:2409.00920},
  year={2024}
}

@inproceedings{srinivasan2023nexusraven,
  title={Nexusraven: a commercially-permissive language model for function calling},
  author={Srinivasan, Venkat Krishna and Dong, Zhen and Zhu, Banghua and Yu, Brian and Mosk-Aoyama, Damon and Keutzer, Kurt and Jiao, Jiantao and Zhang, Jian},
  booktitle={NeurIPS 2023 Foundation Models for Decision Making Workshop},
  year={2023}
}

@article{park2025flexibleefficientgrammarconstraineddecoding,
  title={Flexible and Efficient Grammar-Constrained Decoding},
  author={Park, Kanghee and Zhou, Timothy and D'Antoni, Loris},
  journal={arXiv preprint arXiv:2502.05111},
  year={2025}
}

@article{piotrowski2025lightweightlatentverifiersefficient,
  title={Lightweight Latent Verifiers for Efficient Meta-Generation Strategies},
  author={Piotrowski, Bartosz and Drzewakowski, Witold and Staniszewski, Konrad and Mi{\l}o{\'s}, Piotr},
  journal={arXiv preprint arXiv:2504.16760},
  year={2025}
}

@article{wang2024fantasticsequencesthemfaithful,
  title={FANTAstic SEquences and where to find them: Faithful and efficient API call generation through state-tracked constrained decoding and reranking},
  author={Wang, Zhuoer and Ribeiro, Leonardo FR and Papangelis, Alexandros and Mukherjee, Rohan and Wang, Tzu-Yen and Zhao, Xinyan and Biswas, Arijit and Caverlee, James and Metallinou, Angeliki},
  journal={arXiv preprint arXiv:2407.13945},
  year={2024}
}

@article{patil2023gorillalargelanguagemodel,
  title={Gorilla: Large language model connected with massive apis},
  author={Patil, Shishir G and Zhang, Tianjun and Wang, Xin and Gonzalez, Joseph E},
  journal={Advances in Neural Information Processing Systems},
  volume={37},
  pages={126544--126565},
  year={2024}
}

@article{qwen2.5,
  title={Qwen2. 5 technical report},
  author={Yang, An and Yang, Baosong and Zhang, Beichen and Hui, Binyuan and Zheng, Bo and Yu, Bowen and Li, Chengyuan and Liu, Dayiheng and Huang, Fei and Wei, Haoran and others},
  journal={arXiv preprint arXiv:2412.15115},
  year={2024}
}

@article{openai2024gpt4ocard,
  title={Gpt-4o system card},
  author={Hurst, Aaron and Lerer, Adam and Goucher, Adam P and Perelman, Adam and Ramesh, Aditya and Clark, Aidan and Ostrow, AJ and Welihinda, Akila and Hayes, Alan and Radford, Alec and others},
  journal={arXiv preprint arXiv:2410.21276},
  year={2024}
}

@article{grattafiori2024llama3herdmodels,
  title={The llama 3 herd of models},
  author={Grattafiori, Aaron and Dubey, Abhimanyu and Jauhri, Abhinav and Pandey, Abhinav and Kadian, Abhishek and Al-Dahle, Ahmad and Letman, Aiesha and Mathur, Akhil and Schelten, Alan and Vaughan, Alex and others},
  journal={arXiv preprint arXiv:2407.21783},
  year={2024}
}

@misc{jiang2023mistral7b,
      title={Mistral 7B}, 
      author={Albert Q. Jiang and Alexandre Sablayrolles and Arthur Mensch and Chris Bamford and Devendra Singh Chaplot and Diego de las Casas and Florian Bressand and Gianna Lengyel and Guillaume Lample and Lucile Saulnier and Lélio Renard Lavaud and Marie-Anne Lachaux and Pierre Stock and Teven Le Scao and Thibaut Lavril and Thomas Wang and Timothée Lacroix and William El Sayed},
      year={2023},
      eprint={2310.06825},
      archivePrefix={arXiv},
      primaryClass={cs.CL},
      url={https://arxiv.org/abs/2310.06825}, 
}

@article{li2023api,
  title={Api-bank: A comprehensive benchmark for tool-augmented llms},
  author={Li, Minghao and Zhao, Yingxiu and Yu, Bowen and Song, Feifan and Li, Hangyu and Yu, Haiyang and Li, Zhoujun and Huang, Fei and Li, Yongbin},
  journal={arXiv preprint arXiv:2304.08244},
  year={2023}
}

@article{qu2025tool,
  title={Tool learning with large language models: A survey},
  author={Qu, Changle and Dai, Sunhao and Wei, Xiaochi and Cai, Hengyi and Wang, Shuaiqiang and Yin, Dawei and Xu, Jun and Wen, Ji-Rong},
  journal={Frontiers of Computer Science},
  volume={19},
  number={8},
  pages={198343},
  year={2025},
  publisher={Springer}
}

\clearpage
\appendix



\section{Prompts for Function Call Generation}

\lstdefinestyle{promptstyle}{
    backgroundcolor=\color{gray!5},   
    basicstyle=\ttfamily\small,        
    breaklines=true,                   
    breakatwhitespace=true,            
    captionpos=b,                      
    commentstyle=\color{gray},         
    frame=single,                      
    keepspaces=true,                   
    numbers=left,                      
    numbersep=5pt,                     
    numberstyle=\tiny\color{gray},     
    showspaces=false,                  
    showstringspaces=false,            
    showtabs=false,                    
    tabsize=2,                         
    title=\lstname,                    
    keywordstyle=\color{blue},         
    stringstyle=\color{purple},        
    morekeywords={role, user, system, assistant, content}, 
    emph={Your task is, As an expert, I want you to act as}, 
    emphstyle=\color{teal}\bfseries
}

\begin{lstlisting}[
    style=promptstyle,
    caption={Prompt for generating a function call.}, % 标题
    label={lst:summary_prompt} % 标签，用于正文引用
]<|im_start|>system
You are a helpful assistant.<|im_end|>

<|im_start|>user
[BEGIN OF TASK INSTRUCTION]
You are a tool calling assistant. In order to complete the user's request, you need to select one or more appropriate tools from the following tools and fill in the correct values for the tool parameters. Your specific tasks are:
1. Make one or more function/tool calls to meet the request based on the question.
2. If none of the function can be used, point it out and refuse to answer.
3. If the given question lacks the parameters required by the function, also point it out.

The following are characters that may interact with you
1. user: Provides query or additional information.
2. tool: Returns the results of the tool calling.
[END OF TASK INSTRUCTION]

[BEGIN OF AVAILABLE_TOOLS]
[{"name": "get_weather", "description": "get information about the weather", "parameters": {...}}]
[END OF AVAILABLE_TOOLS]

[BEGIN OF TASK INSTRUCTION]
The output MUST strictly adhere to the following JSON format, and NO other text MUST be included.
The example format is as follows. Please make sure the parameter type is correct. If no function call is needed, please directly output an empty list '[]'

[
{"name": "func_name1", "arguments": {"argument1": "value1", "argument2": "value2"}},
... (more tool calls as required)
]

[END OF TASK INSTRUCTION]

<|im_end|>

<|im_start|>user
Please help me check the weather <|im_end|>

<|im_start|>assistant

\end{lstlisting}

\section{Additional Experimental Results}

\subsection{Can Advanced Reasoning Models Self-Correct Structural Errors?}

An important question is whether stronger reasoning models can implicitly recover from early structural errors through self-reflection, which would challenge our assumption that such errors are difficult to repair once they occur.
To investigate this possibility, we additionally evaluate Qwen2.5-Instruct models with a self-reflection strategy at inference time on BFCL.
Specifically, we compare the original Qwen2.5-7B-Instruct and Qwen2.5-32B-Instruct models against their self-reflection variants, and further contrast them with our Hammer2.1-7B model guided by ToolPRM.

\begin{table}[t]
\centering
\small
\caption{BFCL results of self-reflection baselines. Avg. is the unweighted average over the four splits.}
\resizebox{\columnwidth}{!}{
\setlength{\tabcolsep}{3pt}
\renewcommand\arraystretch{1.03}
\begin{tabular}{@{}lccccc@{}}
\toprule
Model & Simple & Multi. & Paral. & Mul.P. & Avg. \\
\midrule
Qwen2.5-32B & 72.83 & 94.00 & 93.50 & 88.50 & 87.21 \\
Qwen2.5-32B + Refl. & 74.17 & \textbf{95.50} & 93.50 & \textbf{89.00} & 88.04 \\
Qwen2.5-7B & 75.33 & 94.50 & 91.50 & 84.50 & 86.46 \\
Qwen2.5-7B + Refl. & 77.17 & \textbf{95.50} & 91.50 & 83.50 & 86.92 \\
\midrule
\rowcolor{gray!20}
Hammer2.1-7B + ToolPRM & \textbf{79.08} & \textbf{95.50} & \textbf{94.50} & \textbf{89.00} & \textbf{89.52} \\
\bottomrule
\end{tabular}
}
\label{tab:self_reflection_bfcl}
\end{table}

As shown in Table~\ref{tab:self_reflection_bfcl}, self-reflection yields only minor gains for Qwen2.5-Instruct models and still consistently underperforms our 7B model with ToolPRM.
This result suggests that post-hoc reflection is limited for structured outputs such as JSON function calls.
Unlike open-ended reasoning tasks, where a model may revise its line of thought in subsequent sentences, structured generation is inherently brittle: a hallucinated argument value or a subtle early syntax error can invalidate the entire output while remaining difficult for the model itself to detect and repair afterward.
Therefore, these results further support our premise that early errors in structured outputs are functionally unrecoverable in practice, motivating fine-grained intra-call pruning rather than relying on delayed self-correction.

\subsection{Sensitivity to Function Masking}

We further clarify the mechanics and robustness of the function masking strategy used in ToolPRM training.
Function masking is only applied to a subset of the training samples, rather than the entire dataset.
Its role is not merely to prevent memorization of tool names, but also to act as a regularization mechanism that encourages the model to rely on contextual understanding of tool descriptions and argument semantics instead of surface-form cues.

To further validate this design, we conduct an additional ablation study on BFCL by training ToolPRM with and without function masking.
Table~\ref{tab:func_masking_bfcl} shows that removing function masking leads to consistently worse overall performance, and more importantly, weakens the efficacy of inference-time guidance.
These results suggest that partial function masking improves the robustness of ToolPRM and helps it provide more reliable guidance during structured decoding.

\begin{table}[t]
\centering
\small
\caption{Sensitivity of ToolPRM to function masking on BFCL. FM denotes function masking.}
\resizebox{\columnwidth}{!}{
\setlength{\tabcolsep}{3pt}
\renewcommand\arraystretch{1.03}
\begin{tabular}{@{}lccccc@{}}
\toprule
Model & Simple & Multi. & Paral. & Mul.P. & Avg. \\
\midrule
Hammer2.1-1.5B & 74.67 & 92.00 & 84.50 & 80.00 & 82.79 \\
ToolPRM w/o FM & 78.08 & 91.50 & 86.50 & 83.50 & 84.90 \\
\rowcolor{gray!20}
ToolPRM with FM & \textbf{78.42} & \textbf{92.00} & \textbf{87.50} & \textbf{84.50} & \textbf{85.61} \\
\bottomrule
\end{tabular}
}
\label{tab:func_masking_bfcl}
\end{table}

\subsection{Comparison with Constrained Decoding Baselines}

We further compare ToolPRM against constrained decoding baselines by adding FANTASE~\cite{wang2024fantastic}, a state-tracked constrained decoding framework paired with reranking.
For a fair comparison, we equip FANTASE with a RoBERTa reranker that acts as an outcome reward model trained on our constructed ToolPRM dataset.
As shown in Table~\ref{tab:fantase_bfcl}, ToolPRM consistently outperforms the constrained decoding baseline across all splits on BFCL.
This result suggests that enforcing state-level constraints and reranking alone is insufficient to match the benefits of fine-grained process supervision during structured generation.

\begin{table}[t]
\centering
\small
\caption{Comparison between ToolPRM and the constrained decoding baseline FANTASE~\cite{wang2024fantastic} on BFCL.}
\resizebox{\columnwidth}{!}{
\setlength{\tabcolsep}{3pt}
\renewcommand\arraystretch{1.03}
\begin{tabular}{@{}lccccc@{}}
\toprule
Model & Simple & Multi. & Paral. & Mul.P. & Avg. \\
\midrule
Hammer2.1-1.5B & 74.67 & 92.00 & 84.50 & 80.00 & 82.79 \\
+FANTASE & 76.58 & 91.50 & 86.50 & 82.50 & 84.27 \\
\rowcolor{gray!20}
+ToolPRM (Ours) & \textbf{78.42} & \textbf{92.00} & \textbf{87.50} & \textbf{84.50} & \textbf{85.61} \\
\bottomrule
\end{tabular}
}
\label{tab:fantase_bfcl}
\end{table}

\subsection{Generalization Beyond BFCL and ToolAlpaca}

To evaluate whether ToolPRM generalizes beyond BFCL and ToolAlpaca, we conduct additional experiments on the API-Bank benchmark~\cite{li2023api}.
Following prior work on Hammer~\cite{lin2024hammerrobustfunctioncallingondevice}, we use the cleaned version of API-Bank, which contains 314 tool-use dialogues and 753 API calls.
This benchmark evaluates a model's ability to correctly invoke a known API from a query (L1), as well as to select and call APIs from a candidate list (L2).

As shown in Table~\ref{tab:apibank_results}, ToolPRM consistently outperforms the base model and other inference-time baselines on both L1 and L2.
These results validate that the fine-grained supervision learned by ToolPRM generalizes to more complex and out-of-domain tool-use environments, beyond the benchmark settings used in our main experiments.

\begin{table}[t]
\centering
\small
\caption{Generalization results on API-Bank~\cite{li2023api}. Following Hammer~\cite{lin2024hammerrobustfunctioncallingondevice}, we report results on the cleaned test set.}
\resizebox{\columnwidth}{!}{
\setlength{\tabcolsep}{3pt}
\renewcommand\arraystretch{1.03}
\begin{tabular}{@{}lccc@{}}
\toprule
\multicolumn{4}{c}{API-Bank (L1)} \\
\cmidrule(lr){1-4}
Model & F1-API & F1-Args & Avg. \\
\midrule
Hammer2.1-1.5B & 96.55 & 89.96 & 93.26 \\
+Token-level Beam Search & 91.41 & 84.78 & 88.10 \\
+Majority & 93.93 & 85.71 & 89.82 \\
+Best of N & 97.29 & 90.25 & 93.77 \\
\rowcolor{gray!20}
+ToolPRM (Ours) & \textbf{97.82} & \textbf{90.88} & \textbf{94.35} \\
\midrule
\multicolumn{4}{c}{API-Bank (L2)} \\
\cmidrule(lr){1-4}
Model & F1-API & F1-Args & Avg. \\
\midrule
Hammer2.1-1.5B & 85.09 & 67.82 & 76.46 \\
+Token-level Beam Search & 81.50 & 63.84 & 72.67 \\
+Majority & 85.71 & 68.57 & 77.14 \\
+Best of N & 86.98 & 68.49 & 77.74 \\
\rowcolor{gray!20}
+ToolPRM (Ours) & \textbf{87.32} & \textbf{68.89} & \textbf{78.11} \\
\bottomrule
\end{tabular}
}
\label{tab:apibank_results}
\end{table}

\end{document}